\documentclass[letterpaper]{article} 
\usepackage{aaai2027}  
\usepackage[hyphens]{url}  
\usepackage{graphicx} 
\usepackage{natbib}  
\usepackage{caption} 
\usepackage{multirow}
\usepackage{subcaption}
\usepackage{amsmath}
\usepackage{amssymb}

\usepackage[ruled,vlined]{algorithm2e}
\usepackage{algorithmicx}
\usepackage[noend]{algpseudocode}
\usepackage{algorithmicx}
\usepackage[noend]{algpseudocode}

\usepackage{newfloat}
\usepackage{listings}
\DeclareCaptionStyle{ruled}{labelfont=normalfont,labelsep=colon,strut=off} 

\usepackage{booktabs}

\title{Classifier-pruned Bayesian optimization for particle accelerator tuning: Exploring temporally structured manifold of 6D beam phase space}
\author{
    Mahindra Singh Rautela,
    Alan Williams,
    Alexander Scheinker
}
\affiliations{
    Instrumentation \& Controls (AOT-IC/LANSCE), \\
    Los Alamos National Laboratory, 
    Los Alamos, NM, USA \\
    mrautela@lanl.gov
}

\begin{document}

\maketitle

\begin{abstract}
Complex dynamical systems, such as particle accelerators, require tuning over high-dimensional, nonlinear state and parameter spaces while experimental measurements and high-fidelity simulations can be expensive. Learned latent representations provide a compact domain for such optimization, but poorly supported regions of the learned manifold may decode into unrealistic physical states and yield deceptively favorable objectives. To address this challenge, we propose the Classifier-pruned Bayesian Optimization-based Latent-space Tuner (CBOL-Tuner), which performs Bayesian optimization over a temporally structured latent representation of 6D beam phase-space dynamics. The CBOL-Tuner integrates a conditional variational autoencoder for latent space representation, a long short-term memory network for temporal dynamics, a lightweight neural network for parameter estimation, and a classifier-pruned Bayesian optimizer to adaptively search and filter the latent space for optimal solutions. This framework enables feasibility-aware exploration of learned scientific representations for accelerator tuning.
\end{abstract}

\section{Introduction}
\label{sec:intro}
\begin{figure*}[t]
\centering
\includegraphics[trim={0 5mm 0 0},clip, width=0.85\textwidth]{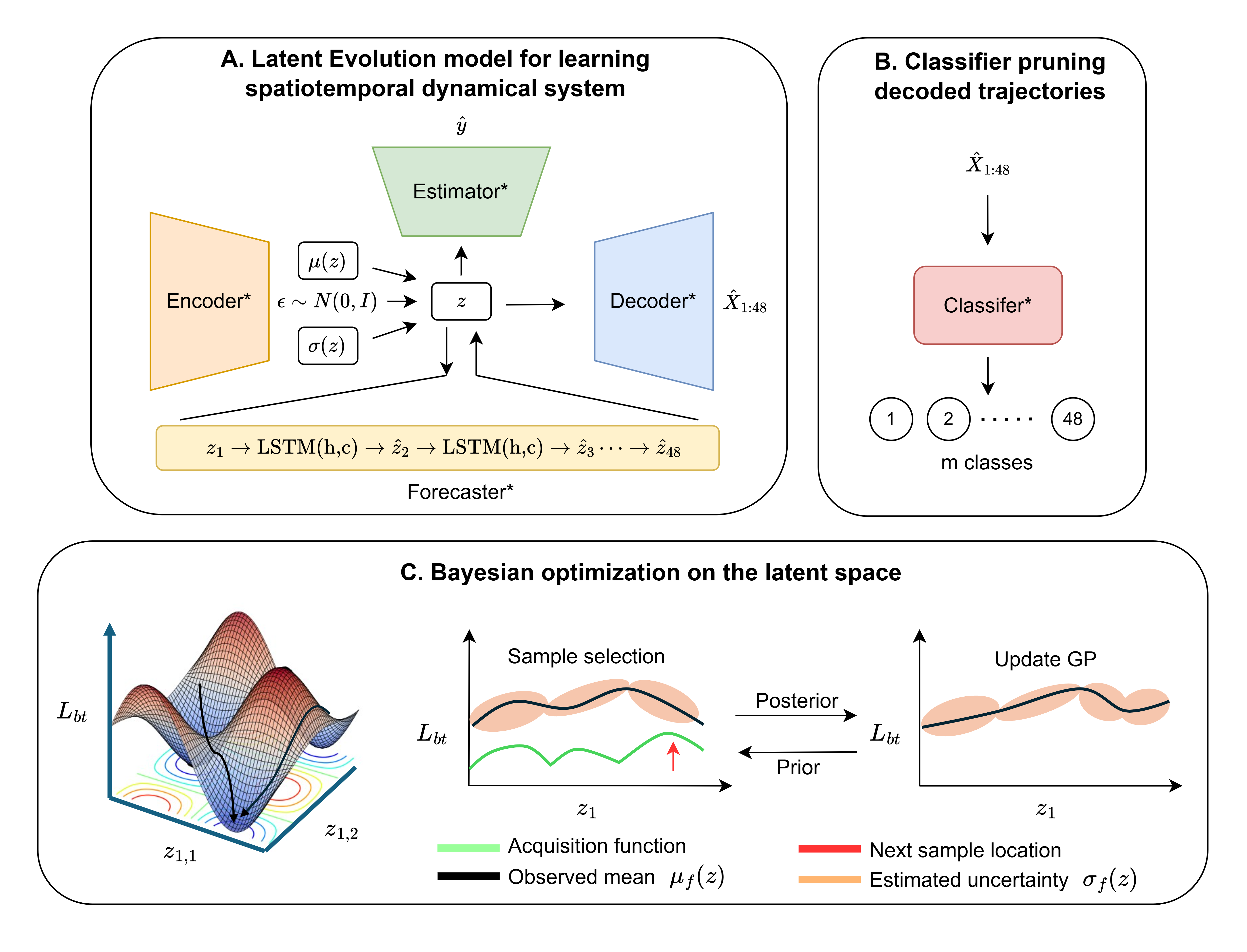}
\caption{Architecture of the CBOL-Tuner for beam optimization in particle accelerators. A. Modular latent evolution model, comprising encoder and decoder of CVAE, forecaster (LSTM) and estimator (DNN) for learning spatiotemporal dynamical system, B. Classifier-pruning using ResNet-50 to classify trajectories into 48 classes (modules), C. Bayesian optimization over the latent space of CVAE. \textbf{(*)} denotes components trained prior to the optimization stage and held fixed during Bayesian optimization..}
\label{fig:model}
\end{figure*}

Complex spatiotemporal dynamical systems evolve through coupled interactions among their states, governing parameters, and initial and boundary conditions, often giving rise to high-dimensional nonlinear dynamics that are expensive to simulate \cite{rautela2025morph}. Learning such dynamics provides efficient surrogate representations for forecasting and, ultimately, optimizing system behavior. Tuning or optimizing the system parameters of complex dynamical systems is a challenging problem, as achieving optimal performance often requires navigating high-dimensional, stochastic, and non-linear parameter spaces. In the context of particle accelerators, the spatiotemporal dynamics of charged particle beams are governed by hundreds to thousands of interconnected components designed to focus, guide, and accelerate intense charged particle beams to high energies \cite{wiedemann1994particle}. During tuning or maintenance periods, before the accelerator is used for scientific applications, system parameters are manually adjusted to optimize performance metrics such as beam loss. However, this process is time-consuming and often leads to suboptimal performance because of the system’s high dimensionality, nonlinear behavior, and susceptibility to distribution shifts \cite{scheinker2023adaptive_PRE}.

From a physics perspective, charged particle dynamics represent a multi-scale spatiotemporal phenomenon, where large ensembles of charged particles interact and evolve under the influence of internal and external electromagnetic forces as they traverse different accelerating sections \cite{scheinker2018demonstration}. The charged particle beam is modeled in a 6D phase space ($x, y, \psi, x', y', E$), whose distribution evolves in a highly nonlinear manner due to collective effects such as space charge forces and coherent synchrotron radiation. On the experimental side, even state-of-the-art experimental characterization can be slow: the first complete 6D beam phase-space measurement at the SNS Beam Test Facility was not single-shot and required repeated measurements over roughly 18 hours \cite{cathey2018first}. In addition, accelerator operation is inherently time-varying, with beam characteristics drifting due to source fluctuations, thermal effects, and changing machine conditions, which can introduce distribution shifts \cite{saxena2026improved,saxena2026mahalanobis,saxenaplanning}. These limitations in measurement, together with the computational cost of high-fidelity physics-based simulations, have driven growing interest in data-driven approaches. In recent years, machine learning has played a significant role in addressing forward, inverse, optimization, and control problems in beam physics \cite{scheinker2024cdvae,edelen2024machine,rautela2025review}.

Bayesian optimization (BO) has emerged as a powerful technique for accelerator tuning, with several studies demonstrating its effectiveness across different particle accelerators \cite{duris2020bayesian,jalas2021bayesian,kirschner2022tuning,hwang2022prior,ji2024multi}. However, the application of Bayesian optimization to tune high-dimensional state and action spaces still remains a significant challenge \cite{moriconi2020high,morita2023accelerator}. Learned latent representations provide a promising alternative for optimization of high-dimensional scientific systems. \emph{By compressing complex physical states into substantially lower-dimensional representations, latent-variable models can provide a computationally tractable domain in which the underlying nonlinear dynamics can be modeled and explored}. This is particularly attractive for accelerator systems, where each state is a  high-dimensional representation of an evolving charged-particle distribution and direct exploration of the original state space is impractical.

However, optimization over a learned latent manifold introduces an important challenge. \emph{Regions of the latent space that are weakly supported by the training data may decode into unrealistic physical states, while appearing favorable with respect to the optimization objective}. An optimizer can therefore exploit imperfections in the learned representation and converge toward solutions that are numerically optimal but physically invalid. This issue becomes especially important in scientific systems where data are sparse, high-fidelity simulations are expensive, and unsafe operating conditions must be avoided.

In this work, we address this problem by introducing the Classifier-pruned Bayesian Optimization-based Latent-space Tuner (CBOL-Tuner), which performs Bayesian optimization over a temporally structured latent representation of the accelerator beam while explicitly screening candidate trajectories for consistency. Rather than optimizing individual beam states independently, the proposed framework searches over an initial latent state whose downstream evolution represents the beam trajectory through the accelerator. Candidate trajectories are decoded into the physical space and filtered using a pretrained classifier before being incorporated into the optimization process. The resulting optimal latent states are subsequently mapped to actionable accelerator control parameters. The key contributions of this work are as follows:

\begin{enumerate}
    \item We formulate accelerator tuning as constrained optimization over a low-dimensional, temporally structured latent manifold of high-dimensional 6D beam phase-space dynamics.

    \item We introduce a classifier-pruned Bayesian optimization strategy in which candidate latent trajectories are decoded and screened for consistency, preventing the optimizer from exploiting poorly represented regions of the learned manifold.

    \item We couple latent-space optimization with inverse parameter estimation, enabling optimized beam trajectories to be translated into actionable accelerator RF settings and subsequently evaluated using a high-fidelity particle-beam simulator.

    \item We demonstrate that latent-space exploration can identify deceptively low-loss states corresponding to nonphysical beam trajectories, highlighting the importance of feasibility-aware optimization in data-limited scientific machine learning.
    
\end{enumerate}

\section{Proposed Method}
\label{sec:methods}
\begin{figure*}[t]
\centering
\includegraphics[trim={0 0 0 0},clip, width=1.0\textwidth]{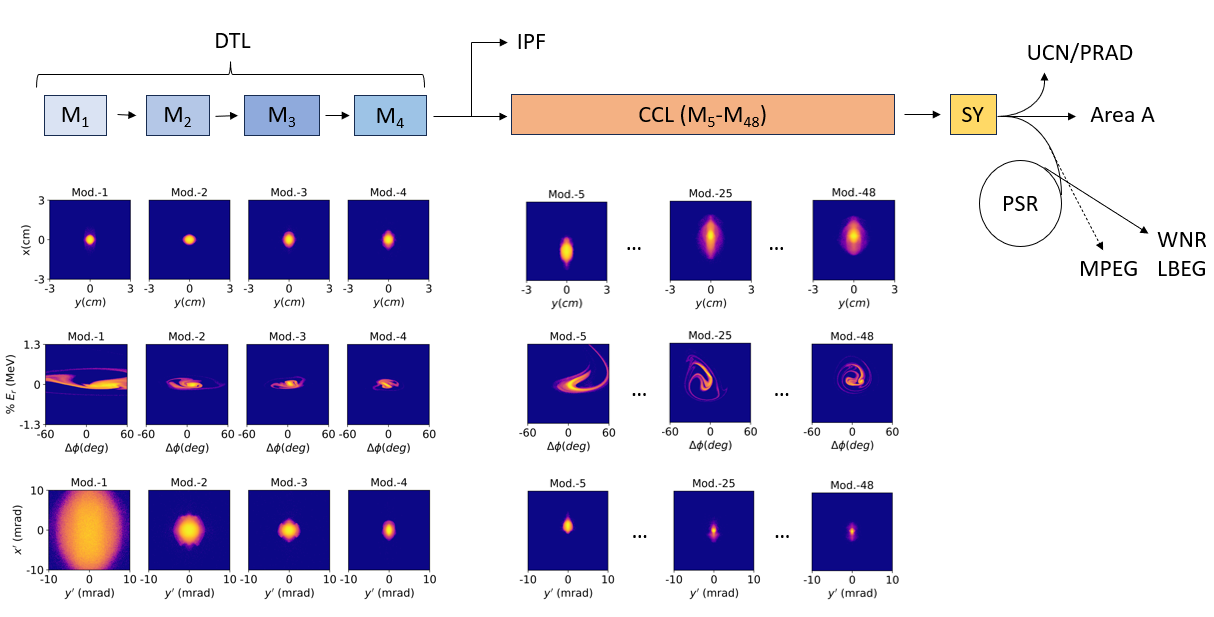}
\caption{3 out of 15 2d projections of 6d phase space of charged particle beam in the LANSCE linear accelerator. Accelerating modules - 1 to 4 are 201 MHz drift tube linac (DTL) and 5 to 48 are 805 MHz coupled cavity linac (CCL). The beam serves various scientific areas like isotope production facility (IPF), ultra-cold neutrons (UCN), proton radiography (PRAD), weapons neutron research (WNR), proton storage ring (PSR).}
\label{fig:dataset}
\end{figure*}

The forward discretized spatiotemporal beam dynamics can be written as 
$X_{t} = H(X_1, X_2,\dots, X_{t-1})$, where $H$ is an unknown nonlinear function, $X_t$ is the high-dimensional state of the system at time $t$. This can be learned as a joint probability distribution over all states, $P(X_1, X_2, \ldots, X_T)$, which can be factorized using the chain rule of probability as
\begin{equation}
\small
P(X_1, X_2, \ldots, X_T) =
P(X_1)\prod_{t=2}^{T} P(X_t \mid X_1, X_2, \ldots, X_{t-1}).
\end{equation}

However, in particle accelerators, $X$ is a high-dimensional object representing 6D phase space comprising positions and momentum of billions of particles. Using variational autoencoders (VAEs), the higher-dimensional distribution can be projected into a lower-dimensional distribution $P(\mathbf{z}_1, \mathbf{z}_2, ..., \mathbf{z}_{T})$ using Bayes' rule $p_{\theta_1}(\mathbf{z}|\mathbf{X}) = p_{\theta_1}(\mathbf{X}|\mathbf{z})p(\mathbf{z})/p_{\theta_1}(\mathbf{X})$. The spatial dynamics is learned by minimizing the Evidence Lower BOund (ELBO) loss:
\begin{eqnarray}
    ELBO(\theta_1,\theta_2;\mathbf{x}) = \mathbb{E}_{\mathbf{z} \sim q_{\theta_2}(\mathbf{z}|\mathbf{x}, m)} [\log p_{\theta_1}(\mathbf{x}|\mathbf{z})] \\
    - D_{KL}(q_{\theta_2}(\mathbf{z}|\mathbf{x},m)||p(\mathbf{z})). \nonumber
\end{eqnarray}

The first term of the loss function captures the reconstruction error between the original and generated data, while the second term involves the Kullback–Leibler (KL) divergence, which quantifies the difference between the approximate and true posterior distributions \cite{kingma2013auto}. 

Temporal evolution of the beam is modeled directly in the latent representation. Given an initial latent state $z_1$, an autoregressive recurrent model generates the downstream latent trajectory
\begin{equation}
    \hat{z}_{t}
    =
    f_{\theta}
    \left(
        \hat{z}_{t-1},
        h_{t-1},
        c_{t-1}
    \right),
    \qquad t=2,\ldots,T,
\end{equation}
where $h_t$ and $c_t$ denote the recurrent hidden and memory states. Consequently, a single candidate $z_1$ induces a temporally structured trajectory $\hat{\mathbf{z}}=\{\hat{z}_1,\ldots,\hat{z}_T\}$ through the accelerator, which is decoded to obtain the corresponding beam evolution
$\hat{\mathbf{X}}=\{\hat{X}_1,\ldots,\hat{X}_T\}$.

The latent representation and temporal evolution model are trained using the simulated beam-dynamics dataset described in the Results section. After training, their parameters are held fixed while the Bayesian optimizer explores the learned latent manifold.

A lightweight inverse mapping relates the learned representation to the accelerator control variables, $\hat{y}=g_{\phi}(z)$, where $y$ denotes the RF settings. Thus, optimization can be performed in the compact latent representation while retaining mappings both to the physical beam state and to actionable machine parameters.

The central objective of this work is therefore the constrained exploration of this learned spatiotemporal manifold for accelerator tuning. We seek a latent state $z^*$ whose induced trajectory minimizes the beam-loss objective while remaining physically consistent. The tuning problem is formulated as

\begin{equation}\label{eq:opt}
    y^*, z^* = \arg \min_{z} L_{bt} = \arg \min_{z} 
    (w(m) \boldsymbol{\cdot} L_b(\hat{X}_m))
\end{equation}

Here, $L_{bt}$ is the total beam loss, which is a weighted sum of beam loss across different modules (or dot product of weighting function and beam loss function), $\hat{X}_m = Decoder(z_m)$ and $(y^*, z^*)$ are the optimal values, where $y^*=Estimator(z^*)$. In our simulations, beam loss is estimated from the transmitted beam current, computed as the total particle count in the 11th projection $(E,\psi)$, where lower transmitted current indicates higher beam loss. Each simulation contains up to approximately $1.04 \times 10^6$ particles.

\begin{algorithm}[t]
    \caption{CBOL-Tuner}
    \label{alg:CBOL-tuner}
    \KwIn{Models/Functions: CVAE, LSTM, DNN, BeamLoss, EI, Classifier; N, $\xi$, $w_{1:48}$}
    \KwOut{Optimized system, $\mathcal{S^*}$: $\{L^*_{bt}, z^*_{1:48}, y^*_{1:8}$\}}
    
    Initialize:
    
    $z_{1} \sim p_{\theta}(z,m=1)$\;     \tcp{Conditional latent sampling}
    $h:z_1 \mapsto L_{bt}$\;                 \tcp{Prior GP}
    
    \For{$iterations \gets 0$ \KwTo N}{
        \tcp{Autoregressive forecasting}
        $\hat{z}_{2:48} \gets LSTM(z_{1})$\; 
        \tcp{Concat}
        $\hat{z}_{1:48} \gets z_{1} \cup \hat{z}_{2:48}$\;
        \tcp{Decoding}
        $\hat{X}_{1:48} \gets Decoder(\hat{z}_{1:48})$\;
        \tcp{Classifier-pruning}
        $\hat{m}_{1:48} \gets Classifier(\hat{X}_{1:48})$\;
        \If{$\hat{m}_{1:48} = m_{1:48}$}
        {
        \tcp{Estimate RF settings}
        $\hat{y}_{1:8} \gets DNN(\hat{z}_{1:48})$\;
        \tcp{Beam Loss}
        $L_{1:48} = BeamLoss(\hat{X}_{1:48})$ \;
        \tcp{Total beam loss}
        $L_{bt} = w_{1:48}  \boldsymbol{\cdot} L_{1:48}$\;
        \tcp{Select new query point}
        $z_1' \gets \arg \max_{z_1 \in Z_1} EI(z_1)$\;
        \tcp{Update GP, i.e, h}
        $h:z_1' \mapsto L_{bt}$\;                             
        $z_1 \gets z_1'$\;
        \tcp{Store results}
        $\mathcal{S} \gets \mathcal{S} \cup \{L_{bt}, z_{1:48}, y_{1:8}\}$\;
        }
    }
    \Return $\mathcal{S}, \mathcal{S^*}$\;
\end{algorithm}

Bayesian Optimization (BO) builds a probabilistic surrogate model, such as a Gaussian process, and uses an acquisition function to balance exploration and exploitation when selecting the next point to evaluate \cite{shahriari2015taking}. We have used expected improvement (EI) as the acquisition function mentioned in Eq.~\ref{eq:ei} \cite{snoek2012practical}. 

\begin{equation}\label{eq:ei}
\begin{aligned}
EI(z) &= 
\begin{cases}
\gamma(z)\Phi(\tilde{z}) + \sigma_f(z)\phi(\tilde{z}), & \text{if } \sigma_f(z) > 0, \\
0, & \text{if } \sigma_f(z) = 0,
\end{cases} \\
\text{where, } \\
\gamma(z) &= \mu_f(z) - f^+ - \xi \text{ ,} \\
\tilde{z} &= \frac{\mu_f(z) - f^+ - \xi}{\sigma_f(z)}. \\
\end{aligned}
\end{equation}

In this context, $z$ represents the design parameter (latent space of the beam). The mean prediction from the Gaussian process at point $z$ is denoted by $\mu_f(z)$, while $\sigma_f^2(z)$ represents the variance (uncertainty) of the prediction at $z$ (See Fig.~\ref{fig:model}(C)).  The best observed value so far is $f^+$. The exploration hyperparameter, $\xi$, controls the exploration-exploitation trade-off during optimization. The standardized improvement is $\tilde{z}$. Here, $\Phi(\tilde{z})$ is the cumulative density function of the standard normal distribution, and $\phi(\tilde{z})$ is its probability density function.

BO operates on the latent space of a CVAE and assumes a well-defined, continuous space for effective exploration. However, achieving an ideal latent space where every randomly sampled point corresponds to realistic physical signals is challenging. To prevent BO from exploring infeasible regions of the latent space, a classifier can assist by either excluding such regions from the explorer's memory or penalizing the optimizer when these regions are encountered. For simplicity, we eliminate infeasible regions from the optimization history using a ResNet50 classifier trained to identify the corresponding accelerator module from reconstructed beam phase-space projections. The classifier achieves an accuracy of approximately $0.99989$ over the 48 accelerator-module classes. Classifier-pruned BO is designed to discard explored points ($z_{1:48}$) if their decoding ($X_{1:48}$) does not belong to the true classes. The CBOL-Tuner algorithm is outlined in Alg.~\ref{alg:CBOL-tuner}.

In the algorithm, $z_1$ is conditionally sampled from the learned latent distribution using a uniform distribution defined by upper and lower bounds for module 1: $z_1 \sim p_{\theta}(z, m=1) \sim U(a, b)$. A Gaussian Process (GP) prior $h$ is initialized to map $z_1$ to the total beam loss. At each iteration, the LSTM predicts $\hat{z}_{2:48}$, which are concatenated with $z_1$ and decoded to obtain $\hat{X}_{1:48}$. The total beam loss is computed, and the acquisition function is maximized to identify the next query point $z_1'$. The GP model is then updated with the new data $h: z_1' \mapsto L_{bt}$. If $\hat{X}_{1:48}$ passes the classifier i.e., all 48 phase spaces are classified correctly to their respective modules (classes), the point is appended to the history $S$. This process is repeated for $N$ iterations. Additionally, the algorithm supports multiple independent runs with the same number of iterations, each initialized with a new $z_1$.

\section{Results}
\label{sec:results}

\begin{figure}
    \centering
    \begin{minipage}[b]{0.49\linewidth}
        \centering
        \includegraphics[width=1.0\textwidth]{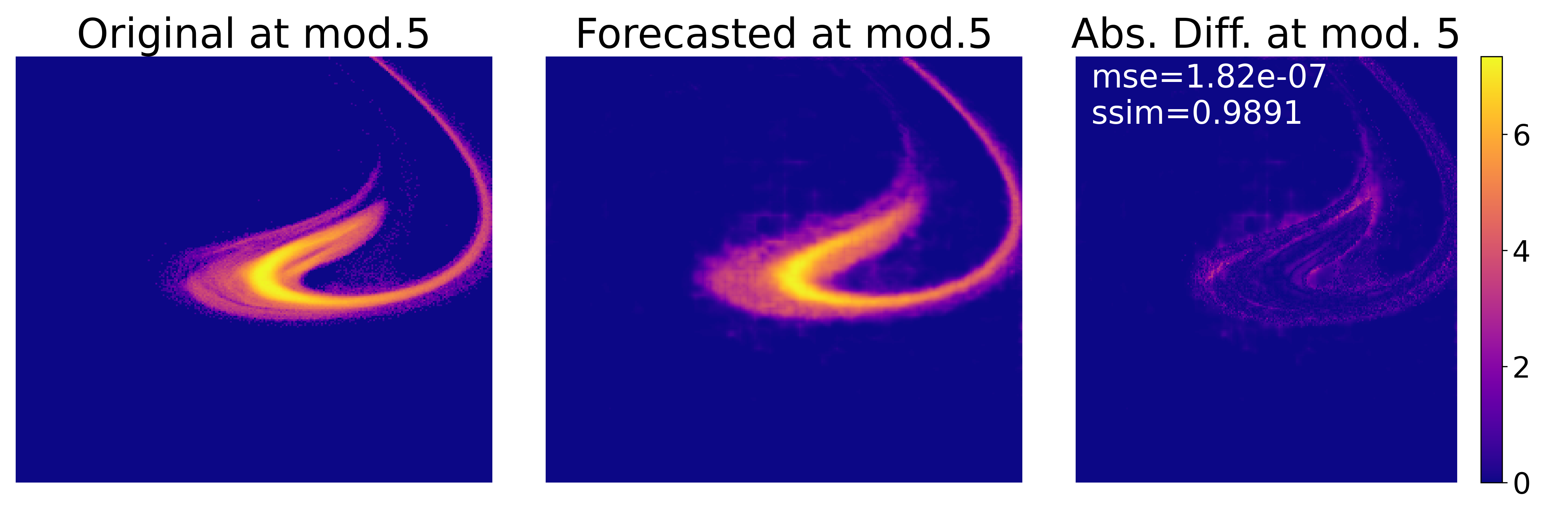}
    \end{minipage}
    \begin{minipage}[b]{0.49\linewidth}
        \centering
        \includegraphics[width=1.0\textwidth]{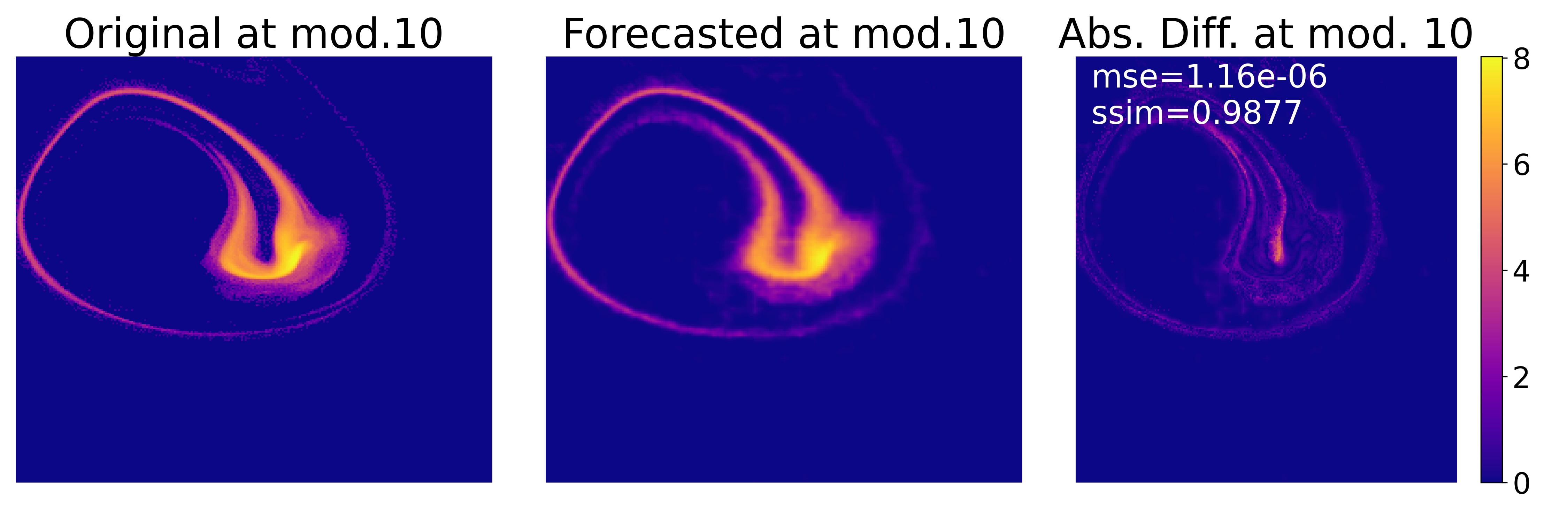}
    \end{minipage}
    \begin{minipage}[b]{0.49\linewidth}
        \centering
        \includegraphics[width=1.0\textwidth]{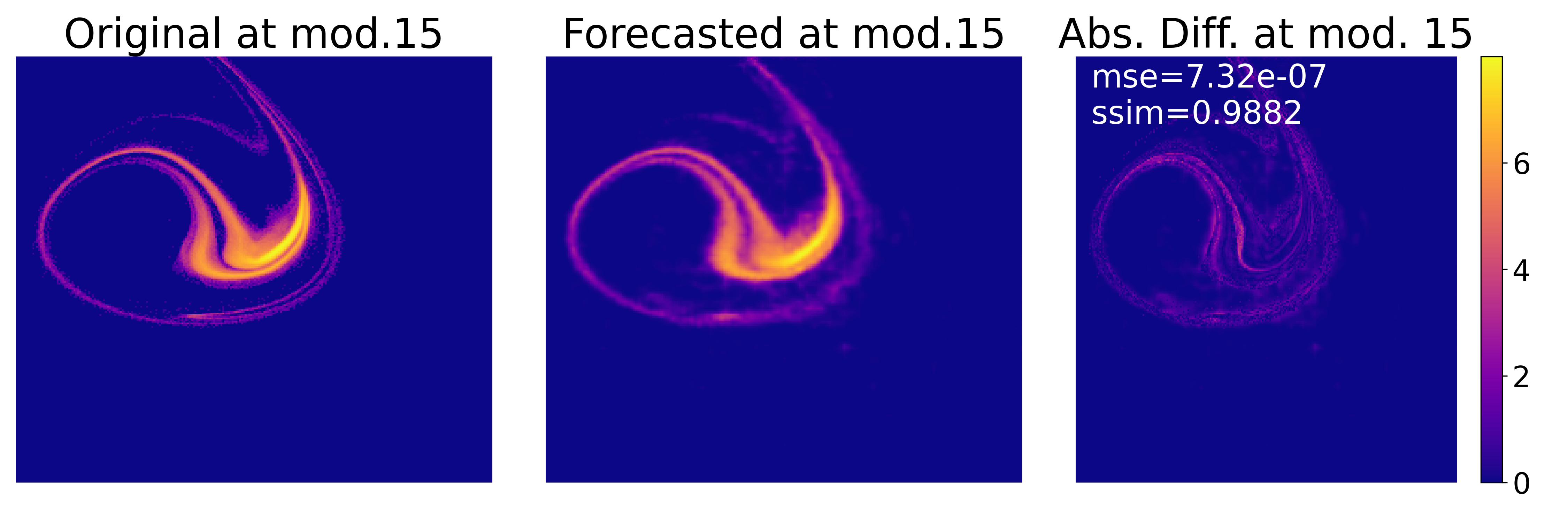}
    \end{minipage}
    \begin{minipage}[b]{0.49\linewidth}
        \centering
        \includegraphics[width=1.0\textwidth]{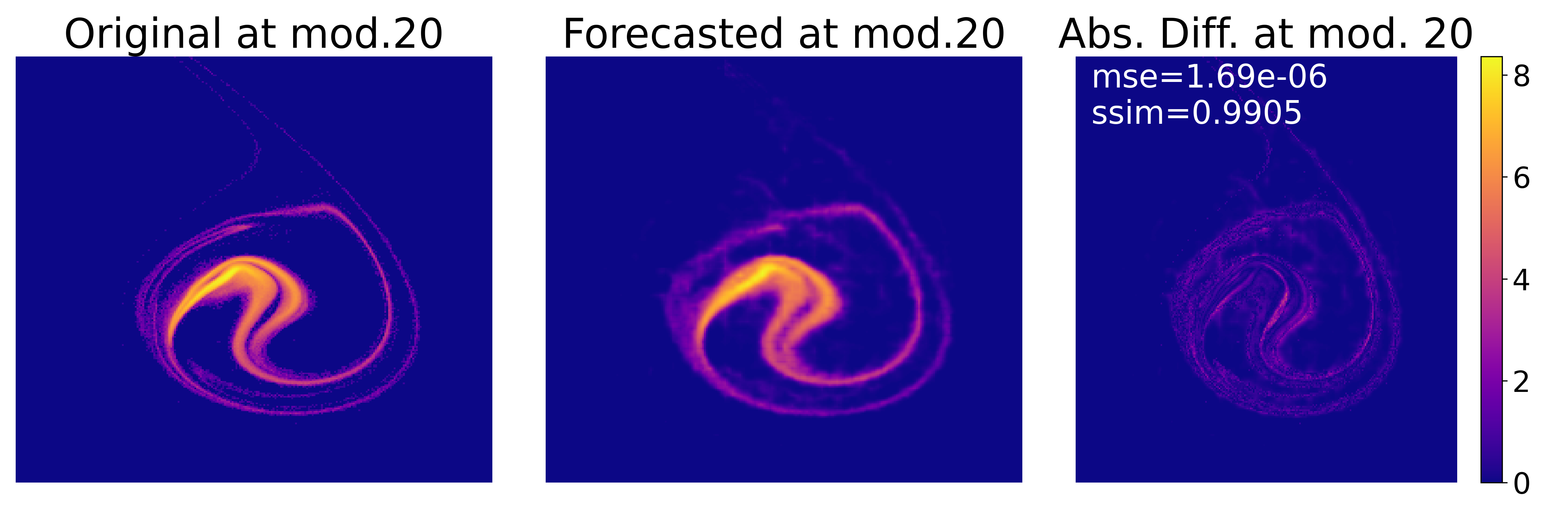}
    \end{minipage}
    \begin{minipage}[b]{0.49\linewidth}
        \centering
        \includegraphics[width=1.0\textwidth]{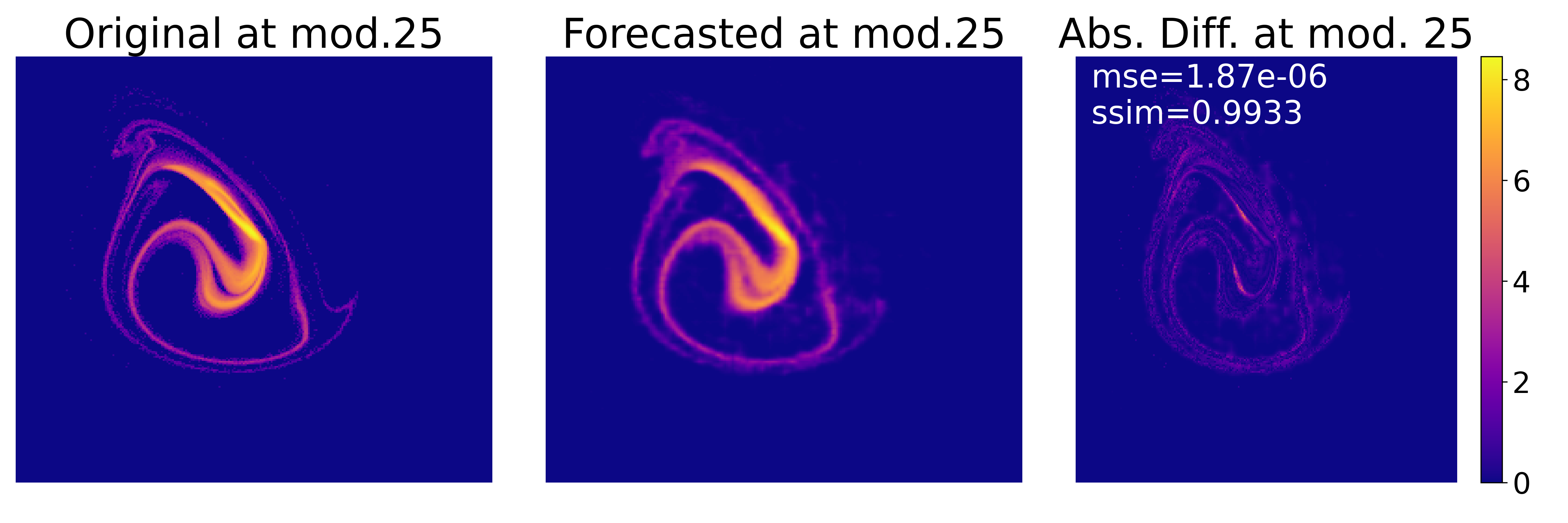}
    \end{minipage}
    \begin{minipage}[b]{0.49\linewidth}
        \centering
            \includegraphics[width=1.0\textwidth]{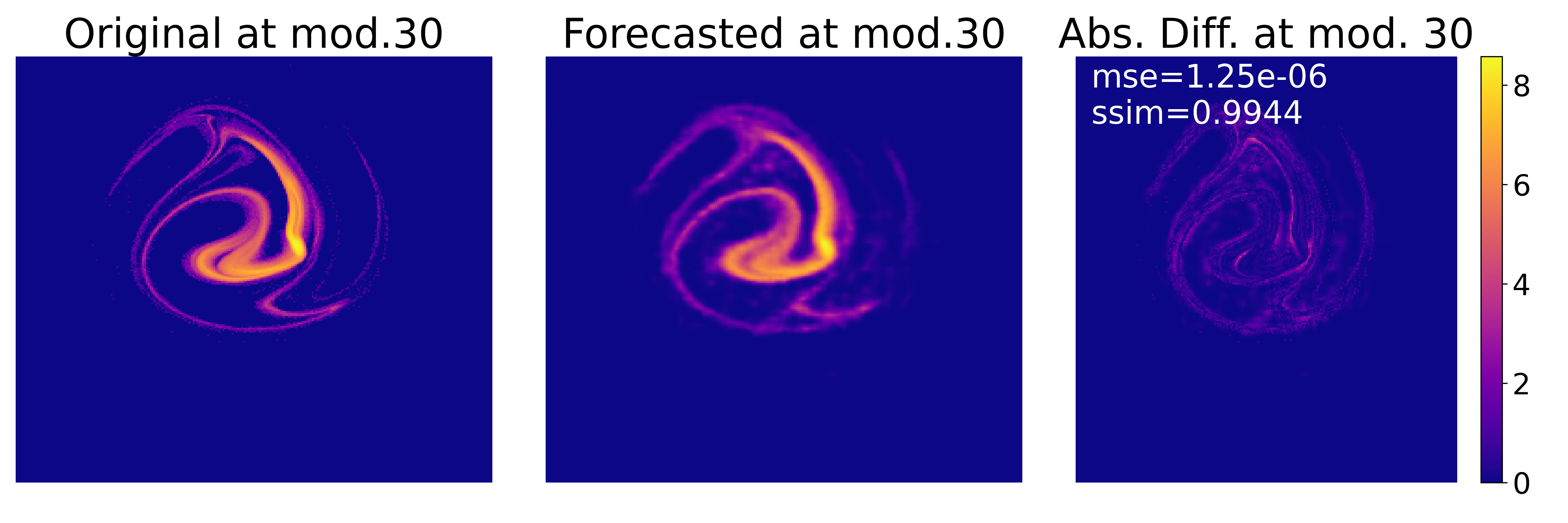}
    \end{minipage}
    \begin{minipage}[b]{0.49\linewidth}
        \centering
        \includegraphics[width=1.0\textwidth]{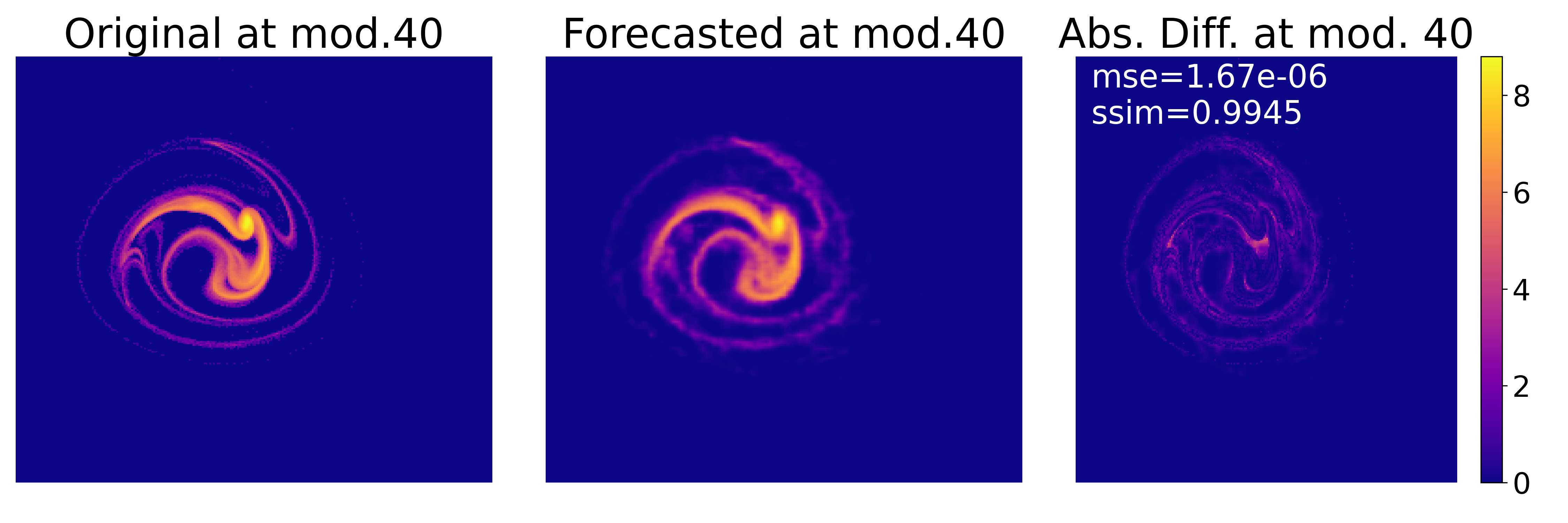}
    \end{minipage}
    \begin{minipage}[b]{0.49\linewidth}
        \centering
        \includegraphics[width=1.0\textwidth]{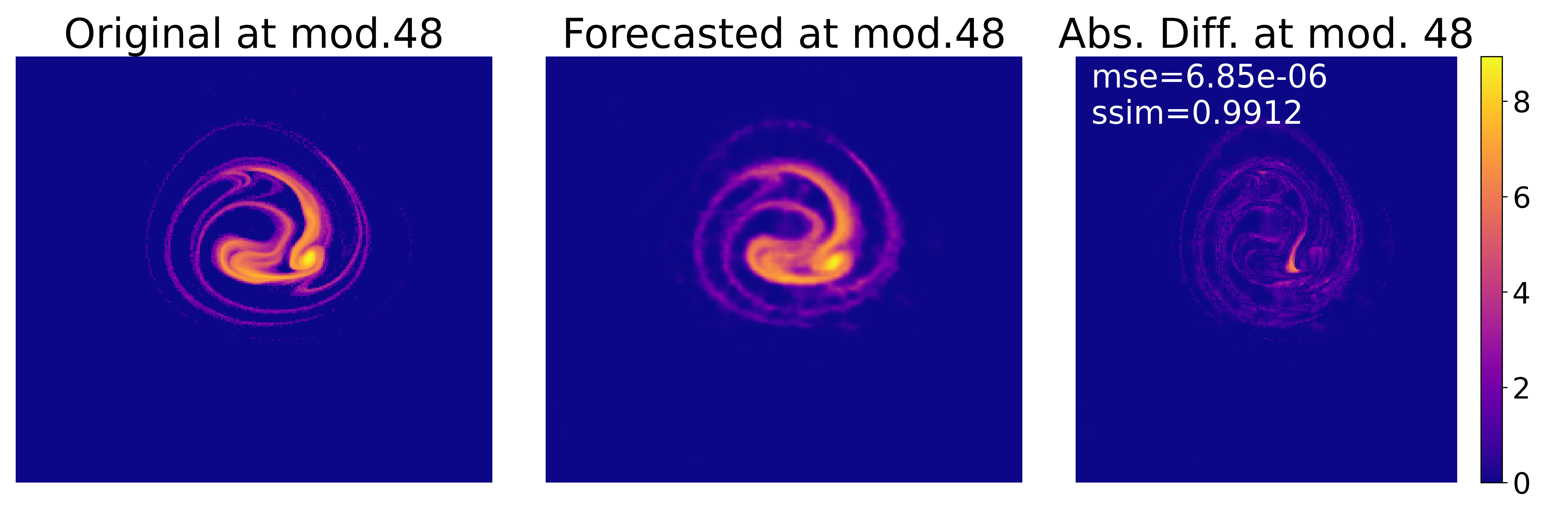}
    \end{minipage}
    \begin{minipage}[b]{1.0\linewidth}
        \centering
        \includegraphics[width=1.0\textwidth]{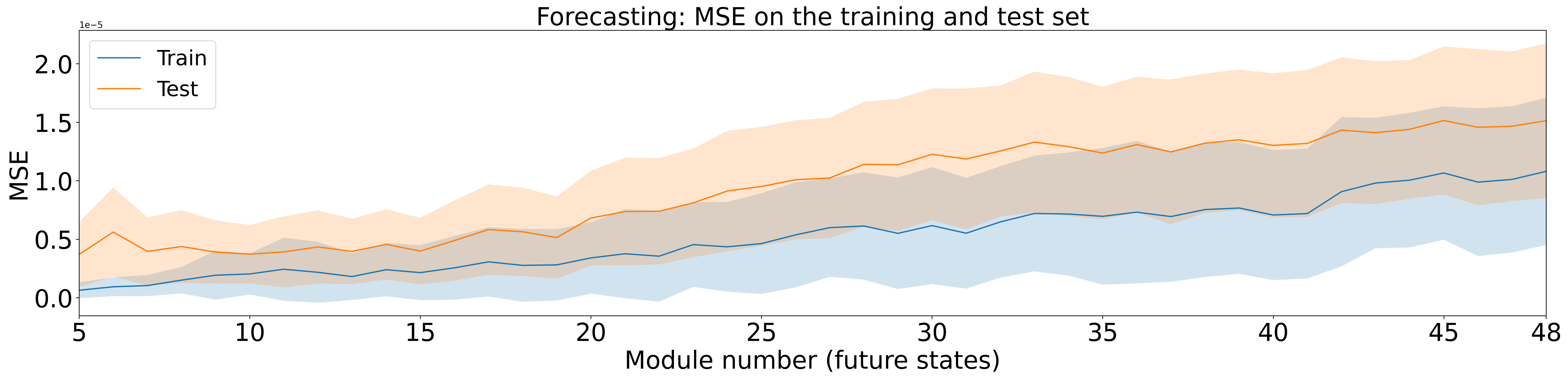}
    \end{minipage}
    \caption{Forecasting results: Forecasted projections (shown $E-\phi$ only) across different modules given first four projections as inputs. The original projection is presented against the forecasted along with the absolute difference between both. The MSE for the entire training and test set are plotted as a shaded plot where the the central line is the mean and the boundaries of the region defines the standard deviation.}
    \label{fig:forecastingability}
\end{figure}

\begin{figure*}[t]
\centering
\includegraphics[trim={0 16mm 0 15mm},clip, width=1.0\textwidth]{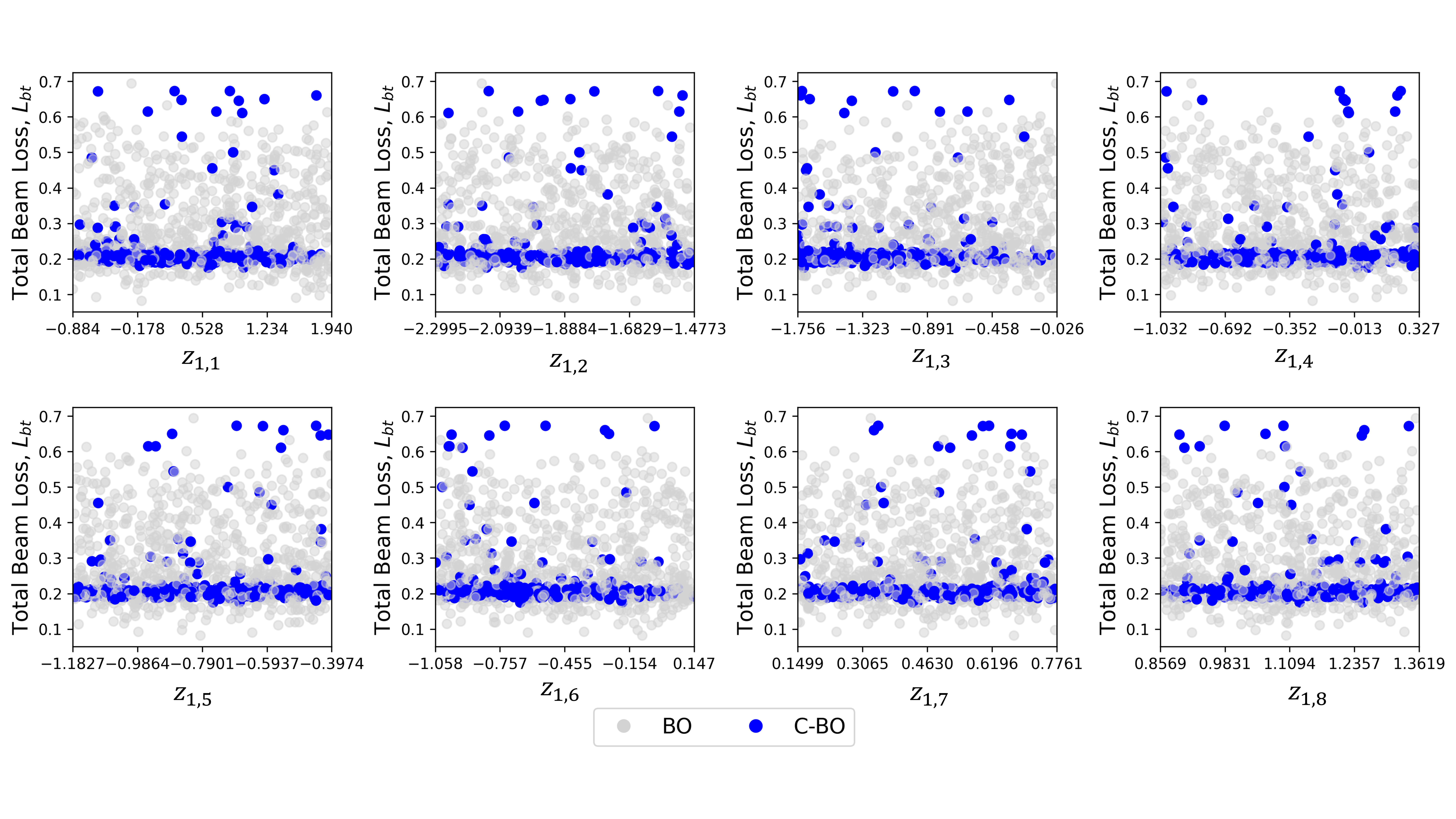}
\caption{Target vs parameter search space for Bayesian optimization and classifier-pruned BO. Classifier-pruning eliminates latent trajectories that do not satisfy the selection criterion.}
\label{fig:bo_vs_cbo}
\end{figure*}

We evaluate CBOL-Tuner on beam dynamics of the Los Alamos Neutron Science Center (LANSCE) linear accelerator. The underlying charged-particle dynamics constitute a nonlinear, high-dimensional system governed by self-consistent particle--field interactions. We use the High-Performance Simulator (HPSim), a GPU-accelerated multi-particle beam-dynamics simulator developed at Los Alamos National Laboratory~\cite{pang2014gpu}, to generate the data used for training and evaluation.

High-Performance Simulator (\texttt{HPSim}) is an open-source, GPU-accelerated code developed at Los Alamos National Laboratory (LANL) for simulations of multi-particle beam dynamics \cite{pang2014gpu}. The software is designed to replicate the accelerator and, therefore, provides a realistic representation of the true beam used at the Los Alamos Neutron Science Center (LANSCE). The beam behavior in each module is manipulated by two RF settings, i.e., amplitude and phase set-points. The beam passes through 48 modules, including 4 modules of a 201.25 MHz drift tube linac and 44 modules of an 805 MHz coupled cavity linac.

We generated the dataset by randomly sampling the RF set points of the first four modules (i.e., 8 settings) from a uniform distribution with $\pm 1\%$ variation bounds, while keeping the RF set points of the remaining 44 modules (88 settings) constant. The simulator outputs a 6D beam, which is subsequently projected into 2D to obtain 15 unique projections for each of the 48 modules. A total of 1400 simulations were performed to generate the training data, and an additional 100 simulations for testing, resulting in tensor shape: $[48, 15, 256, 256]$ for every simulation. The module numbers ($m$) and projections ($X$) were normalized to the range $[0, 1]$, while the RF settings ($y$) were normalized to $[-0.5, 0.5]$. An example showcasing three projections across the 48 modules is presented in Fig.~\ref{fig:dataset}. The figure highlights the complex, nonlinear, and multi-scale spatiotemporal evolution observed in the projections of the 6D beam phase space.

The CBOL-Tuner (Fig.~\ref{fig:model} and Alg.~\ref{alg:CBOL-tuner}) is designed to search for optimal latent points that minimize the total beam loss in the accelerator (or maximize negative of the beam loss). The total beam loss is computed as the dot product of the weighting vector \(w \in \mathbb{R}^{48}\) and the beam loss vector \(L_b \in \mathbb{R}^{48}\). Since downstream scientific applications prioritize beam quality at the accelerator exit, the operating objective is concentrated on the final module \cite{williams2024experimental}. Therefore, in this work, we define \(w\) as a step function with zero weights for the first 47 modules and unit weight for the final module. 

We first train an eight-dimensional latent representation of the beam dynamics together with the temporal evolution, inverse parameter mapping, and trajectory classification models using the HPSim dataset described above \cite{rautela2024conditional,rautela24accelerator}. These models define the learned spatiotemporal representation used by CBOL-Tuner. To assess forecasting accuracy, Fig.~\ref{fig:forecastingability} shows the $E-\phi$ projection using $M_1$--$M_4$ as inputs (see Supplementary Figs.~18--19 for $x-y$ and $x'-y'$). The forecasts closely match the ground truth, with MSE on the order of $10^{-7}$ and SSIM above 0.99. Errors are initially concentrated near the distribution tails and increase downstream, reflected by increasing MSE while SSIM remains nearly constant. This trend is expected from the autoregressive rollout: after predicting $\hat{M}_5$, the LSTM recursively uses its own predictions, causing CVAE and LSTM errors to accumulate across modules.

Following training, Bayesian optimization is performed over the initial latent state $z_1\in\mathbb{R}^{8}$. Each proposed $z_1$ is propagated through the 48 accelerator modules, decoded into the corresponding beam trajectory, and screened by the classifier before being admitted to the feasible search history. The optimization is performed for 10 independent runs, each containing $N=1000$ iterations, using Expected Improvement with $\xi=0.1$.

Figure~\ref{fig:bo_vs_cbo} illustrates the BO search space for a single run. The figure shows 8-dimensional $z_1$ and the corresponding values of the objective function, \(L_{bt}\). The search space is highly nonlinear, non-convex, and high-dimensional. The x-ticks in each of the eight subplots represent the bounds of \(z_1\), which constrain BO’s exploration within the feasible set. As shown in the figure (gray dots), BO successfully explores the entire 8-dimensional latent space. Blue dots represent the subset of \(z_1\) points that, when forecasted to \(\hat{z}_{2:48}\) and decoded to \(\hat{X}_{1:48}\), yield trajectories classified as valid by the ResNet50 classifier. Notably, many latent space points corresponding to the lowest beam loss, \(L_{bt}\), are rejected by the classifier for failing to represent realistic beam states. This rejection stems from the limited size of the training dataset, which results in a latent space that lacks perfect continuity and smoothness everywhere. This sparsity leads to zones in the latent space that generate unrealistic or hallucinated beam states, some of which may yield low beam loss values.

To further assess CBOL-Tuner, we conduct five additional independent runs, each consisting of 1{,}000 iterations, and validate the resulting solutions using \texttt{HPSim}. For each run, the optimizer identifies an optimal latent point \(z^*\), which is decoded to produce a predicted optimal total beam loss \(L_{bt}^*\). The predicted losses are \([0.1721, 0.1710, 0.1718, 0.1638, 0.1673]\). The associated optimal RF settings \(y^*\), obtained from the DNN using \(z^*\) as input, are then applied in \texttt{HPSim} to generate the corresponding 6D beam and compute the simulator-based total beam losses, which are \([0.2070, 0.2197, 0.2193, 0.2186, 0.2187]\). Although an offset exists between the predicted and simulator-evaluated losses, the optimization procedure remains consistent across runs, producing significantly lower values of total beam loss. This discrepancy highlights a limitation of the learned surrogate representation and suggests that improved agreement may be achieved with a larger and more representative training dataset.

A notable advantage of the proposed methodology is its ability to generate multiple optimal RF configurations, which is particularly beneficial for tuning particle accelerators. With additional runs, the CBOL-Tuner can identify a broader range of such optimal configurations, further enhancing its utility in accelerator tuning applications. 

\paragraph{Future work.}
Currently, the optimization framework employs total beam loss as the sole objective within a single-objective optimization setup. While this approach has proven effective, it has the flexibility to accommodate other important parameters of beam dynamics that are essential for beam optimization. Additional objectives such as minimizing beam emittance, controlling its growth, reducing energy spread, enforcing parameter-space constraints and optimizing beam size, can be incorporated into the framework. Incorporating these objectives would extend the problem into a multi-objective optimization problem, demanding Pareto front analysis to effectively navigate and balance trade-offs among competing objectives. Additionally, we aim to enhance the performance of the CVAE, LSTM and DNN models by incorporating a larger and more diverse dataset, which is expected to improve model generalization and predictive capability.

\section{Conclusions}
\label{sec:conclusions}

In this paper, we propose the classifier-pruned Bayesian Optimization-based Latent-space Tuner (CBOL-Tuner) for constrained exploration of temporally structured scientific representations. Applied to particle-accelerator tuning, the framework searches a compact latent description of 6D beam phase-space dynamics while screening decoded trajectories for physical consistency. This feasibility mechanism prevents the optimizer from favoring poorly represented latent regions that can produce deceptively low-loss but nonphysical beam states. Feasible latent optima are mapped to accelerator RF settings and evaluated with a high-fidelity particle-beam simulator, providing an end-to-end path from learned beam representations to actionable machine configurations. The results highlight the potential of feasibility-aware latent-space optimization for data-limited and computationally expensive scientific systems.

\section*{Acknowledgments}
Los Alamos National Laboratory (LANL) is operated by Triad National Security, LLC, for the national Nuclear Security Administration of the US Department of Energy (Contract No. 89233218CNA000001). This work was supported by the LANL LDRD Program Directed Research (DR) project 20220074DR.

\bibliography{main}

\end{document}